\documentclass[11pt,a4paper]{article}
\usepackage[hyperref]{acl2019}
\usepackage{times}
\usepackage{latexsym}

\usepackage{multirow}
\usepackage{cleveref}
\usepackage{graphicx}
\usepackage{colortbl}
\usepackage{enumerate}
\usepackage{booktabs} 
\usepackage{amsmath}
\definecolor{mygray}{gray}{.9}
\Crefformat{section}{\S#2#1#3}
\usepackage{CJK}

\usepackage{bm}

\usepackage{color}
\usepackage[textsize=scriptsize]{todonotes}
\definecolor{blue}{RGB}{0, 93, 170}			%Go Big Blue!
\definecolor{darkgreen}{RGB}{0, 102, 0}

\usepackage{url}

\aclfinalcopy % Uncomment this line for the final submission
\def\aclpaperid{659} %  Enter the acl Paper ID here

\title{Cross-lingual Knowledge Graph Alignment via Graph~Matching~Neural~Network}

\author{Kun Xu$^{1}$, Liwei Wang$^{1}$, Mo Yu$^{2}$, Yansong Feng$^{3}$, Yan Song$^{1}$, Zhiguo Wang$^{4}$, Dong Yu$^{1}$\\
         $^{1}$Tencent AI Lab \\ 
	 $^{2}$IBM T.J. Watson Research \\
	 $^{3}$Peking University\\
	 $^{4}$Amazon AWS\\
	\{\texttt{syxu828},\texttt{wlwsjtu1989},\texttt{zgw.tomorrow}\}\texttt{@gmail.com}\\
	\texttt{yum}\texttt{@us.ibm.com}, \texttt{fengyansong}\texttt{@pku.edu.cn},
	\{\texttt{clksong,dyu}\}\texttt{@tencent.com}\\
}

\date{}

\begin{document}
\maketitle
\begin{CJK*}{UTF8}{gbsn}

\begin{abstract}
Previous cross-lingual knowledge graph (KG) alignment studies rely on entity embeddings derived
only from monolingual KG structural information, which may fail at matching entities that have different
facts in two KGs.
In this paper, we introduce the \emph{topic entity graph}, a local sub-graph of an entity, to represent 
entities with their contextual information in KG.
From this view, the KB-alignment task can be formulated as a graph matching problem; and we further propose a graph-attention based solution, which first matches all entities in two topic entity graphs, and then 
jointly model the local matching information to derive a graph-level matching vector.
Experiments show that our model outperforms previous state-of-the-art methods by a large margin.
% and view the KB-alignment task as a graph matching problem.
% To achieve this, we further propose a graph matching model which first matches all entities in two topic entity graphs, and then 
% jointly model these local matching information to derive a graph matching vector.
% Experimental results show that our model outperforms previous state-of-the-art methods by a large margin.
% \mo{Previous cross-lingual knowledge graph (KG) alignment works rely on entity embeddings derived from
% the monolingual KG structural information. Contextual information of entities is thus not efficiently utilized.}
% the monolingual KG structural information, which may fail at matching entities that have many different neighbors in two KGs.
% \mo{The problem is not quite clear to me. Seems just lack of context information}
% the KG contextual information of an entity
% and view this task as a graph matching problem.
% \mo{
% In this paper, we defined the \emph{topic entity graph}, a local sub-graph of an entity, to represent its contextual information in KG. In this way, the KB-alignment task can be naturally viewed as a graph matching problem.} 
\end{abstract}

\section{Introduction}
Multilingual knowledge graphs (KGs), such as DBpedia \cite{DBLP:dblp_conf/semweb/AuerBKLCI07} and Yago \cite{DBLP:conf/www/SuchanekKW07}, represent human knowledge in the structured format and have been successfully used in many natural language processing applications.
These KGs encode rich monolingual knowledge but lack the cross-lingual links to bridge the language gap.
Therefore, the cross-lingual KG alignment task, which automatically matches entities in a multilingual KG, is proposed to address this problem.

\begin{figure}[t!]
\centering\includegraphics[width=0.45\textwidth]{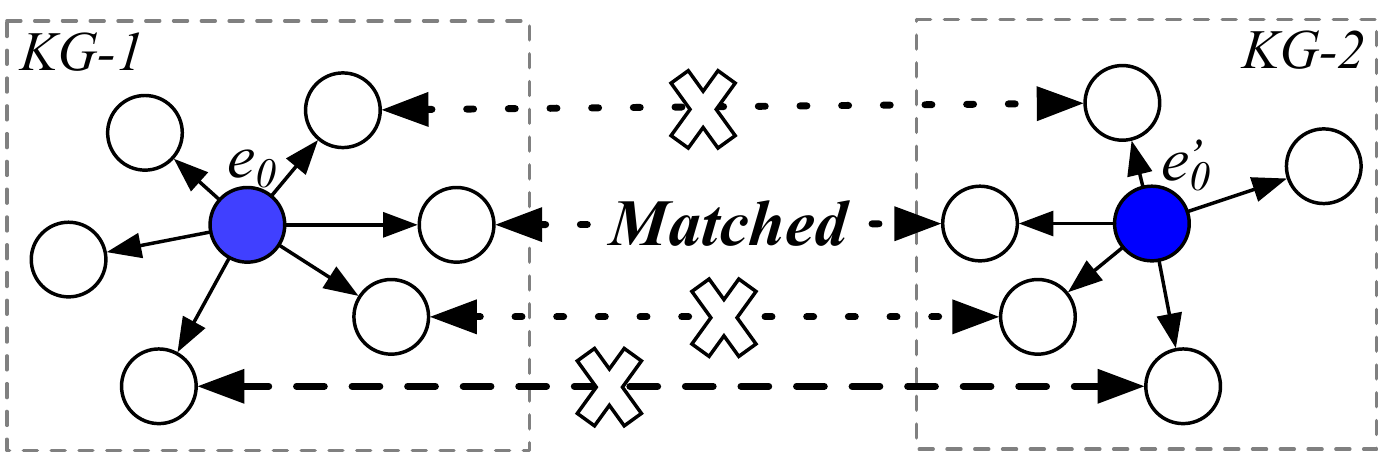}
\caption{A challenging entity matching example.}
\vspace{-4mm}
\label{fig:intro_example}
\end{figure}

Most recently, several entity matching based approaches \cite{hao2016joint,chen2016multilingual,sun2017cross,wang2018cross} have been proposed for this task.
% \mo{Essentially, the aligned entities in different languages may have different facts, making the information encoded in entity embeddings diverse cross-languages.}
Generally, these approaches first project entities of each KG into low-dimensional vector spaces by encoding monolingual KG facts, and then learn a similarity score function to match entities based on their vector representations.
However, since some entities in different languages may have different KG facts, the information encoded in entity embeddings may be diverse across languages, making it difficult for these approaches to match these entities.
% Since these approaches derive the entity embeddings using only monolingual KG facts, they may fail to match entities that have very different KG context information.
Figure~\ref{fig:intro_example} illustrates such an example where
we aim to align $e_{0}$ with $e_{0}'$, but there is only one aligned neighbor in their surrounding neighbors.
% which makes it challenging for existing methods to correctly match them.
In addition, these methods do not encode the entity surface form into the entity embedding, also making it difficult to match entities that have few neighbors in the KG that lacks sufficient structural information.

% Intuitively, both the entity surface form and its KG structure information should be encoded into its embedding.

To address these drawbacks, we propose a \textit{topic entity graph} to represent the KG context information of an entity.
Unlike previous methods that utilize entity embeddings to match entities, we formulate this task as a graph matching problem between the topic entity graphs.
To achieve this, we propose a novel graph matching method to estimate the similarity of two graphs.
Specifically, we first utilize a graph convolutional neural network (GCN) \cite{kipf2016semi,hamilton2017inductive} to encode two graphs, say $G_{1}$ and $G_{2}$,
resulting in a list of entity embeddings for each graph. Then, we compare each entity in $G_{1}$ (or $G_{2}$) against all entities in $G_{2}$ (or $G_{1}$)
by using an attentive-matching method, which generates cross-lingual KG-aware matching vectors for all entities in $G_{1}$ and $G_{2}$.
\begin{figure*}[ht!]
\centering\includegraphics[width=1.0\textwidth]{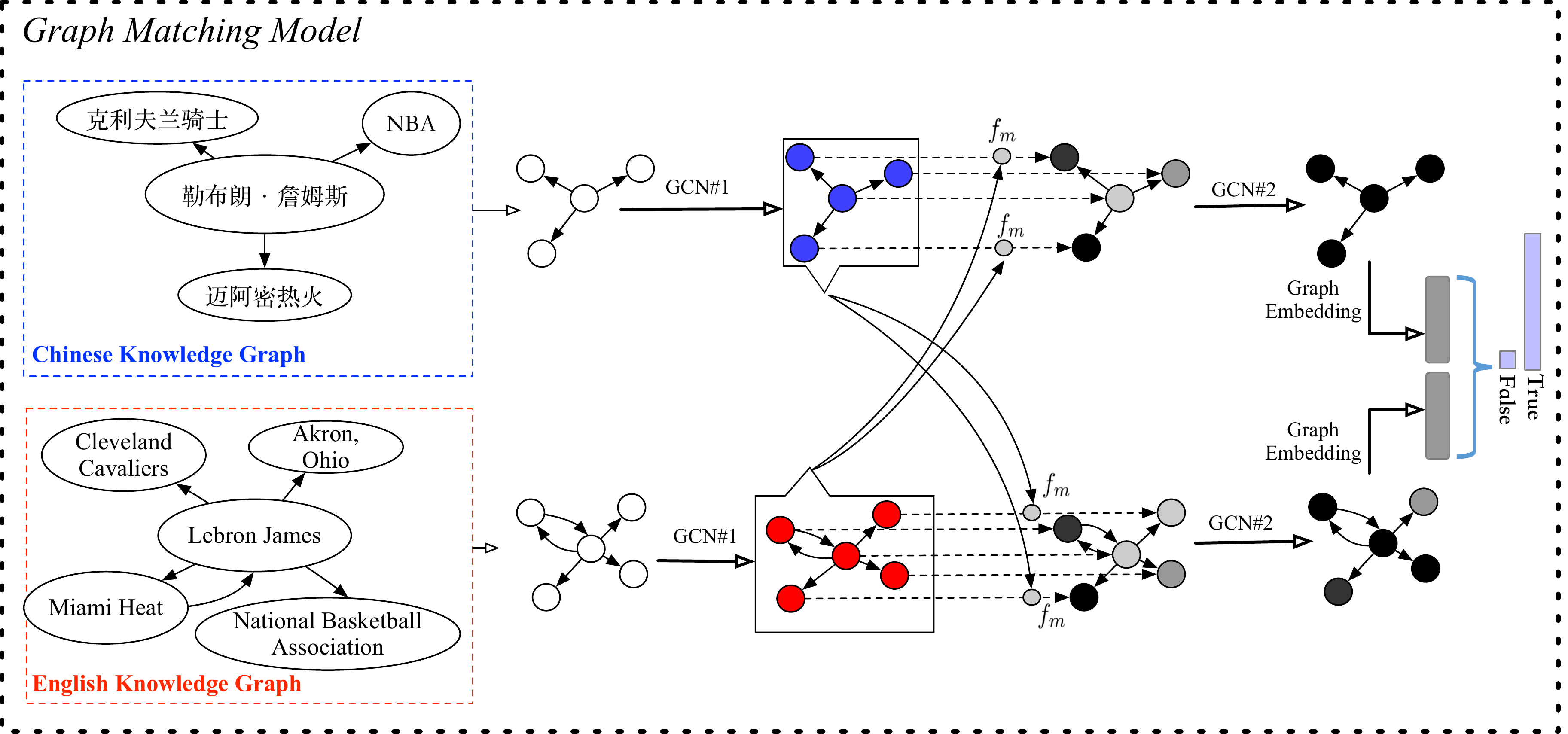}
\caption{A running example of our model for aligning \textit{Lebron James} in the English and Chinese knowledge graph.}
% \vspace{-4mm}
\label{fig:model}
\end{figure*}
Consequently, we apply another GCN to propagate the local matching information throughout the entire graph.
This produces a global matching vector for each topic graph that is used for the final prediction.
% \mo{The contribution of graph-encoding can be actually merged with the new problem formulation. Another major contribution is graph matching network (GMN), which contains graph-based attention and aggregation that have not been studies before.}
The motivation behind is that, the graph convolution could jointly encode all entity similarities, including both the topic entity and its neighbor entities, into a matching vector.
% Finally, we feed two global matching vectors to a prediction layer.
Experimental results show that our model outperforms previous state-of-the-art models by a large margin.
% Both our graph matching framework and the incoperation of entity surface form significantly contribute to this large improvement.
Our code and data is available at {\small \url{https://github.com/syxu828/Crosslingula-KG-Matching}}.

\section{Topic Entity Graph}
% \mo{Is this an unlabeled directed graph containing the target entity and its one-hot neighbors? The edge relation does not work because we did not match the relations (the edges can be attended and matched as well) ?}
As indicated in \newcite{wang2018cross}, the local contextual information of an entity in the KG is important to the KG alignment task. In our model, we propose a structure, namely \textit{topic entity graph}, to represent relations among the given entity (called \textit{topic entity}) and its neighbors in the knowledge base.
Figure~\ref{fig:model} shows the topic graphs of \textit{Lebron James} in the English and Chinese knowledge graph.
In order to build the topic graph, we first
collect $1$-hop neighbor entities of the topic entity, resulting in a set of entities, \{$e_{1}$, ..., $e_{n}$\},
which are the nodes of the graph.
Then, for each entity pair ($e_{i}$, $e_{j}$), we add one directed edge between their corresponding nodes in the topic graph if $e_{i}$ and $e_{j}$
are directly connected through a relation, say $r$, in the KG. Notice that, we do not label this edge with $r$ that $e_{i}$ and $e_{j}$ hold in the KG,
but just retain $r$'s direction. In practice, we find this strategy significantly improves both the efficiency and performance, which we will discuss in \Cref{exp_sec}.

\section{Graph Matching Model}
Figure~\ref{fig:model} gives an overview of our method for aligning \textit{Lebron James} in the English and Chinese knowledge graph\footnote{Lebron James is translated to {\small 勒布朗·詹姆斯} in Chinese.}.
Specifically, we fist retrieve topic entity graphs of \textit{Lebron James} from two KGs, namely $G_{1}$ and $G_{2}$.
Then, we propose a graph matching model to estimate the probability that $G_{1}$ and $G_{2}$ are describing the same entity.
In particular, the matching model includes the following four layers:

\paragraph{Input Representation Layer}
The goal of this layer is to learn embeddings for entities that occurred in topic entity graphs by using a GCN (henceforth \textit{GCN}$_{1}$) \cite{xu2018graph2seq}.
Recently, GCN has been successfully applied in many NLP tasks, such as semantic parsing \cite{xu2018exploiting}, text representation \cite{zhang2018sentence}, relation extraction \cite{song2018n} and text generation \cite{xu2018sql}. 
We use the following embedding generation of entity $v$ as an example to explain the GCN algorithm:
\\
\textbf{(1)}
We first employ a word-based LSTM to transform $v$'s entity name to its initial feature vector \textbf{a}$_{v}$;\\
\textbf{(2)} We categorize the neighbors of $v$ into incoming neighbors $\mathcal{N}_{\vdash}(v)$ and outgoing neighbors $\mathcal{N}_{\dashv}(v)$ according to the edge direction.\\
% Specifically, $\mathcal{N}_{\vdash}(v)$ returns the nodes that $v$ directs to and $\mathcal{N}_{\dashv}(v)$ returns the nodes that direct to $v$; 
\textbf{(3)} We leverage an aggregator to aggregate the \textbf{incoming} representations of $v$'s incoming neighbors \{\textbf{h}$_{u\vdash}^{k-1}$, $\forall u \in \mathcal{N}_{\vdash}(v)$\} into a single vector, \textbf{h}$_{\mathcal{N}_{\vdash}(v)}^{k}$, where $k$ is the iteration index.
This aggregator feeds each neighbor's vector to a fully-connected neural network and applies an element-wise mean-pooling operation to capture different aspects of the neighbor set.
% Notice that, at iteration $k$, this aggregator only uses the representations generated at $k-1$. The initial forward representation of each node is its feature vector calculated in step (1); 
\\
\textbf{(4)} We concatenate $v$'s current \textbf{incoming} representation \textbf{h}$_{v\vdash}^{k-1}$ with the newly generated neighborhood vector \textbf{h}$_{\mathcal{N}_{\vdash}(v)}^{k}$
and feed the concatenated vector into a fully-connected layer to update the \textbf{incoming} representation of $v$, \textbf{h}$_{v\vdash}^{k}$ for the next iteration; \\
\textbf{(5)} We update the \textbf{outgoing} representation of $v$, \textbf{h}$_{v\dashv}^{k}$ using the similar procedure as introduced in step (3) and (4) except that operating on the outgoing representations;\\
\textbf{(6)} We repeat steps (3)$\sim$(5) by $K$ times and treat the concatenation of final incoming and outgoing representations as the final representation of $v$.
The outputs of this layer are two sets of entity embeddings $\{\textit{\textbf{e$^{1}_{1}$}},...,\textit{\textbf{e$^{1}_{|G_{1}|}$}}\}$ and $\{\textit{\textbf{e$^{2}_{1}$}},...,\textit{\textbf{e$^{2}_{|G_{2}|}$}}\}$.
% Since the neighbor information from different hops may have different impact on the node embedding, we learn a distinct aggregator at each iteration.

\paragraph{Node-Level (Local) Matching Layer}
In this layer, we compare each entity embedding of one topic entity graph against all entity embeddings of the other graph in both ways (from $G_{1}$ to $G_{2}$ and from $G_{2}$ to $G_{1}$), as shown in Figure~\ref{fig:model}.
% we match each entity of $G_{1}$ against all entities of $G_{2}$, and match each entity of $G_{2}$ against all entities of $G_{1}$.
% To match one entity against all entities of the other graph, 
We propose an attentive-matching method similar to \cite{wang2017bilateral}.
Specifically, we first calculate the cosine similarities of entity $e_{i}^{1}$ in $G_{1}$ with all entities $\{e_{j}^{2}\}$ in $G_{2}$ in their representation space.
\begin{displaymath}
\begin{split}
\alpha_{i,j}=cosine(\bm{e}^{1}_{i},\bm{e}_{j}^{2}) \hspace{10mm} j \in {\{1,...,|G_{2}|\}} \nonumber 
\label{equ:attentions}
\end{split}
\end{displaymath}
Then, we take these similarities as the weights to calculate an attentive vector for the entire graph $G_{2}$ by weighted summing all the entity embeddings of $G_{2}$.
\vspace{-3mm}
\begin{displaymath}
\begin{split}
\bar{\bm{e}}_{i}^{1}= \frac{\sum_{j=1}^{|G_{2}|} \alpha_{i,j} \cdot \bm{e}_j^2}{\sum_{j=1}^{|G_{2}|} {\alpha}_{i,j}} \nonumber \\
\label{equ:weighted_sum}
\end{split}
\end{displaymath}
\vspace{-8mm}

\noindent We calculate matching vectors for all entities in both $G_{1}$ and $G_{2}$
by using a multi-perspective cosine matching function $f_m$ at each matching step (See Appendix~\ref{matching_layer} for more details):
% \vspace{-2.5mm}
\begin{equation}
\begin{split}
\bm{m}^{att}_i = f_{m}(\bm{e}_i^1,\bar{\bm{e}}_i^1) \nonumber \\
\bm{m}^{att}_j = f_{m}(\bm{e}_j^2,\bar{\bm{e}}_j^2) \nonumber \\
\label{equ:attentive_matching}
\end{split}
\end{equation}
% \vspace{-8mm}

\paragraph{Graph-Level (Global) Matching Layer}
Intuitively, the above matching vectors ($\bm{m}^{att}$s) capture how each entity in $G_{1}$ ~($G_{2}$) can be matched by the topic graph in the other language. 
However, they are \textbf{\textit{local}} matching states and are not sufficient to measure the \textbf{\textit{global}} graph similarity.
For example, many entities only have few neighbor entities that co-occurr in $G_{1}$ and $G_{2}$.
For those entities, a model that exploits local matching information may have a high probability to incorrectly
predict these two graphs are describing different topic entities since most entities in $G_{1}$ and $G_{2}$ are not close in their embedding space.

To overcome this issue, we apply another GCN (henceforth \textit{GCN}$_{2}$) to propagate the local matching information throughout the graph.
Intuitively, if each node is represented as its own matching state, by design a GCN over the graph (with a sufficient number of hops) is able to encode the global matching state between the pairs of whole graphs.
We then feed these matching representations to a fully-connected neural network and apply the element-wise \textit{max} and \textit{mean} pooling method to generate a fixed-length graph matching representation.

\paragraph{Prediction Layer}
We use a two-layer feed-forward neural network to consume the fixed-length graph matching representation and apply the \textit{softmax} function in the output layer.

\paragraph{Training and Inference}
To train the model, we randomly construct $20$ negative examples for each positive example $<$$e_{i}^{1}, e_{j}^{2}$$>$ using a heuristic method. That is, we first generate rough entity embeddings for $G_{1}$ and $G_{2}$ by summing over the pretrained embeddings of words within each entity's surface form; then, we select $10$ closest entities to $e_{i}^{1}$ (or $e_{j}^{2}$) in the rough embedding space to construct negative pairs with $e_{j}^{2}$ (or $e_{i}^{1}$).
During testing, given an entity in $G_{1}$, we rank all entities in $G_{2}$ by the descending order of matching probabilities that estimated by our model.

\section{Experiments}
\label{exp_sec}
\begin{table*}[t!]
\centering
\small
\begin{tabular}{l|cc|cc|cc|cc|cc|cc}
\toprule[0.8pt]
\multirow{2}{*}{Method} & \multicolumn{2}{c|}{\textit{ZH}-\textit{EN}} & \multicolumn{2}{c|}{\textit{EN}-\textit{ZH}} & \multicolumn{2}{c|}{\textit{JA}-\textit{EN}} & \multicolumn{2}{c|}{\textit{EN}-\textit{JA}} & \multicolumn{2}{c|}{\textit{FR}-\textit{EN}} & \multicolumn{2}{c}{\textit{EN}-\textit{FR}} \\
\cline{2-13}
& @1 & @10 & @1 & @10 & @1 & @10 & @1 & @10 & @1 & @10 & @1 & @10 \\
\hline
% \newcite{hao2016joint}
Hao (2016)& 21.27 & 42.77 & 19.52 & 39.36 & 18.92 & 39.97 & 17.80 & 38.44 & 15.38 & 38.84 & 14.61 & 37.25 \\
% \newcite{chen2016multilingual} 
Chen (2016)& 30.83 & 61.41 & 24.78 & 52.42 & 27.86 & 57.45 & 23.72 & 49.92 & 24.41 & 55.55 & 21.26 & 50.60  \\
% \newcite{sun2017cross} 
Sun (2017)& 41.18 & 74.46 & 40.15 & 71.05 & 36.25 & 68.50 & 38.37 & 67.27 & 32.39 & 66.68 & 32.97 & 65.91 \\
% \newcite{wang2018cross} 
Wang (2018)& 41.25 & 74.38 & 36.49 & 69.94 & 39.91 & 74.46 & 38.42 & 71.81 & 37.29 & 74.49 & 36.77 & 73.06 \\
BASELINE & 59.64 & 72.30 & 57.66 & 70.44 & 67.01 & 79.53 & 62.48 & 77.54 & 83.45 & 91.56 & 81.03 & 90.79 \\
\hline
\hline
\textit{NodeMatching} & 62.03 & 75.12 & 60.17 & 72.67 & 69.82 & 80.19 & 66.74 & 80.10 & 84.71 & 92.35 & 84.15 & 91.76  \\
% \textit{GraphMatching} 
\textbf{Ours} & & & & & & & & & & & &\\
Hop$_{GCN_{2}}$ = 1 & 66.91 & 77.52 & 64.01 & 78.12 & 72.63 & 85.09 & 69.76 & 83.48 & 87.62 & 94.19 & 87.65 & 93.66 \\
% \quad 
Hop$_{GCN_{2}}$ = 3 
&  \textbf{67.93} & \textbf{78.48} & \textbf{65.28} & \textbf{79.64} & \textbf{73.97} & \textbf{87.15} & \textbf{71.29} & \textbf{84.63} & \textbf{89.38} & \textbf{95.24} & \textbf{88.18} & \textbf{94.75}\\
% \quad 
Hop$_{GCN_{2}}$ = 5 
& 67.92 & 78.36 & 65.21 & 79.48 & 73.52 & 86.87 & 70.18 & 84.29 & 88.96 & 94.28 & 88.01 & 94.37 \\
\toprule[0.8pt]
\end{tabular}
\caption{Evaluation results on the datasets.}
\label{tab:results}
\end{table*}

We evaluate our model on the DBP15K datasets, which were built by \newcite{sun2017cross}.
The datasets were generated by linking entities in the Chinese, Japanese and French versions of DBpedia into English version.
Each dataset contains 15,000 inter-language links connecting equivalent entities in two KGs of different languages. We use the same train/test split as previous works.
We use the Adam optimizer \cite{DBLP:journals/corr/KingmaB14} to update parameters with mini-batch size 32.
The learning rate is set to 0.001.
The hop size $K$ of \textit{GCN$_{1}$} and \textit{GCN$_{2}$} are set to 2 and 3, respectively.
The non-linearity function $\sigma$ is ReLU \citep{DBLP:journals/jmlr/GlorotBB11} and
the parameters of aggregators are randomly initialized.
Since KGs are represented in different languages, we first retrieve monolingual fastText embeddings \cite{bojanowski2017enriching} for each language, and apply the method
proposed in \newcite{conneau2017word} to align these word embeddings into a same vector space, namely,
cross-lingual word embeddings. We use these embeddings to initialize word representations in the first layer of \textit{GCN$_{1}$}.
\paragraph{Results and Discussion.}
Following previous works, we used \textit{Hits}@1 and \textit{Hits}@10 to evaluate our model, where
\textit{Hits}@\textit{k} measures the proportion of correctly aligned entities ranked in the top $k$.
We implemented a baseline (referred as BASELINE in Table~\ref{tab:results}) that selects $k$ closest $G_{2}$ entities to a given $G_{1}$ entity in the cross-lingual embedding space, where an entity embedding is the sum of embeddings of words within its surface form.
We also report results of an ablation of our model (referred as \textit{NodeMatching} in Table~\ref{tab:results}) that uses \textit{GCN$_1$} to derive the two topic entity embeddings and then directly feeds them to the prediction layer without using matching layer.
Table~\ref{tab:results} summarizes the results of our model and existing works.

We can see that even without considering any KG structural information, the BASELINE significantly outperforms previous works that mainly learn entity embeddings from the KG structure, indicating that the surface form is an important feature for the KG alignment task. Also, the \textit{NodeMatching}, which additionally encodes the KG structural information into entity embeddings using \textit{GCN$_1$}, achieves better performance compared to the BASELINE.
In addition, we find the graph matching method significantly outperforms all baselines, which suggests that the global context information of topic entities is important to establish their similarities.

Let us first look at the impacts of hop size of \textit{GCN$_{2}$} to our model.
From Table~\ref{tab:results}, we can see that our model could benefit from increasing the hop size of \textit{GCN$_{2}$} until
it reaches a threshold $\lambda$.
In experiments, we find the model achieves the best performance when $\lambda$ = 3.
To better understand on which type of entities that our model could better deal with due to introducing the graph matching layer, we analyze the entities that our model correctly predicts while \textit{NodeMatching} does not. We find the graph matching layer enhances the ability of our model in handling the entities whose most neighbors in two KGs are different.
For such entities, although most local matching information indicate that these two entities are \textit{irrelevant},
the graph matching layer could alleviate this by propagating the most relevant local matching information throughout the graph.

Recall that our proposed topic entity graph only retains the relation direction while neglecting the relation label.
In experiments, we find incorporating relation labels as distinct nodes that connecting entity nodes into the topic graph hurts not only the performance but efficiency of our model.
We think this is due to that (1) relation labels are represented as abstract symbols in the datasets, which provides quite limited knowledge about the relations, making it difficult for the model to learn their alignments in two KGs;
(2) incorporating relation labels may significantly increase the topic entity graph size, which requires bigger hop size and running time.

\section{Conclusions}
Previous cross-lingual knowledge graph alignment methods mainly rely on entity embeddings that derived from the monolingual KG structural information, thereby may fail at matching entities that have different facts in two KGs.
To address this, we introduce the topic entity graph to represent the contextual information
of an entity within the KG and view this task as a graph matching problem.
For this purpose, we further propose a graph matching model which induces a graph matching vector by jointly encoding the entity-wise matching information.
Experimental results on the benchmark datasets show that our model significantly outperforms existing baselines.
In the future, we will explore more applications of the proposed idea of attentive graph matching.
For example, the metric learning based few-shot knowledge base completion~\cite{xiong2018one} can be directly formulated as a similar graph matching problem in this paper.

\bibliography{acl2018}
\bibliographystyle{acl_natbib}
% \clearpage
\appendix

\section{Matching Function $f_m$}
\label{matching_layer}
$f_m$ is a multi-perspective cosine matching function that compares two vectors
\begin{equation}
\bm{m} = f_{m}(\bm{v}_1,\bm{v}_2;\bm{W}) \nonumber
\label{equ:MP_cosine}
\end{equation}
where $\bm{v}_1$ and $\bm{v}_2$ are two $d$-dimensional vectors, $\bm{W} \in \Re^{l \times d}$ is a trainable parameter with the shape $l \times d$, $l$ is the number of perspectives, and the returned value $\bm{m}$ is a $l$-dimensional vector $\bm{m}=[m_1,...,m_k,...,m_l]$. Each element $m_k \in \bm{m}$ is a matching value from the $k$-th perspective, and it is calculated by the cosine similarity between two weighted vectors
\begin{equation}
m_k=cosine(W_k \circ \bm{v}_1, W_k \circ \bm{v}_2) \nonumber
\label{equ:weight_cosine}
\end{equation}
where $\circ$ is the element-wise multiplication, and $W_k$ is the $k$-th row of $\bm{W}$, which controls the $k$-th perspective and assigns different weights to different dimensions of the $d$-dimensional space.

%\bibliography{acl2018}

\end{CJK*}
\end{document}